\documentclass[10pt,twocolumn,letterpaper]{article}

\usepackage[pagenumbers]{cvpr} % To force page numbers, e.g. for an arXiv version

\usepackage{graphicx}
\usepackage{amsmath, mathrsfs}
\usepackage{amssymb}
\usepackage{booktabs}
\usepackage{multirow}
\usepackage{enumitem}
\usepackage[pagebackref,breaklinks,colorlinks]{hyperref}

\usepackage[capitalize]{cleveref}
\crefname{section}{Sec.}{Secs.}
\Crefname{section}{Section}{Sections}
\Crefname{table}{Table}{Tables}
\crefname{table}{Tab.}{Tabs.}

\def\cvprPaperID{10224} % *** Enter the CVPR Paper ID here
\def\confName{CVPR}
\def\confYear{2023}

\begin{document}

\title{DCT Perceptron Layer: A Transform Domain Approach for Convolution Layer}

\author{Hongyi Pan\\
		\and
		Xin Zhu\\
		\and
		Salih Atici
		\and
		Ahmet Enis Cetin\\
		\and
		Department of Electrical and Computer Engineering\\
		University of Illinois Chicago\\
		{\tt\small \{hpan21, xzhu61, satici2, aecyy\}@uic.edu}\\
	}
\maketitle

%%%%%%%%% ABSTRACT
\begin{abstract}
In this paper, we propose a novel Discrete Cosine Transform (DCT)-based neural network layer which we call DCT-perceptron to replace the $3\times3$ Conv2D layers in the Residual neural Network (ResNet). Convolutional filtering operations are performed in the DCT domain using element-wise multiplications by taking advantage of the Fourier and DCT Convolution theorems. A trainable soft-thresholding layer is used as the nonlinearity in the DCT perceptron. Compared to ResNet's Conv2D layer which is spatial-agnostic and channel-specific, the proposed layer is location-specific and channel-specific. The DCT-perceptron layer reduces the number of parameters and multiplications significantly while maintaining comparable accuracy results of regular ResNets in CIFAR-10 and ImageNet-1K. Moreover, the DCT-perceptron layer can be inserted with a batch normalization layer before the global average pooling layer in the conventional ResNets as an additional layer to improve classification accuracy.
\end{abstract}

%%%%%%%%% BODY TEXT
\section{Introduction}
\label{sec:intro}
%GG%Recent literature states that 
Convolutional neural networks (CNNs) have produced remarkable results in image classification~\cite{krizhevsky2012imagenet,simonyan2014very, szegedy2015going, he2016identity, wang2017residual, he2016deep, badawi2020computationally, pan2020computationally, agarwal2021coronet, partaourides2020self, stamoulis2018designing, liu2022convnet}, object detection~\cite{redmon2016you, aslan2020deep, menchetti2019pain, aslan2019early, liu2022learning} and semantic segmentation~\cite{yu2018bisenet, huang2019ccnet, long2015fully, poudel2019fast, jin2019fast}. One of the most widely used and successful CNNs is ResNet~\cite{he2016deep}, which can build very deep networks. However, the number of parameters significantly increases as the network goes deeper to obtain better performance in accuracy. However, the huge amount of parameters increases the computational load for the devices, especially for edge devices with limited computational resources.

\begin{figure}
    \centering
    \includegraphics[width=0.8\linewidth]{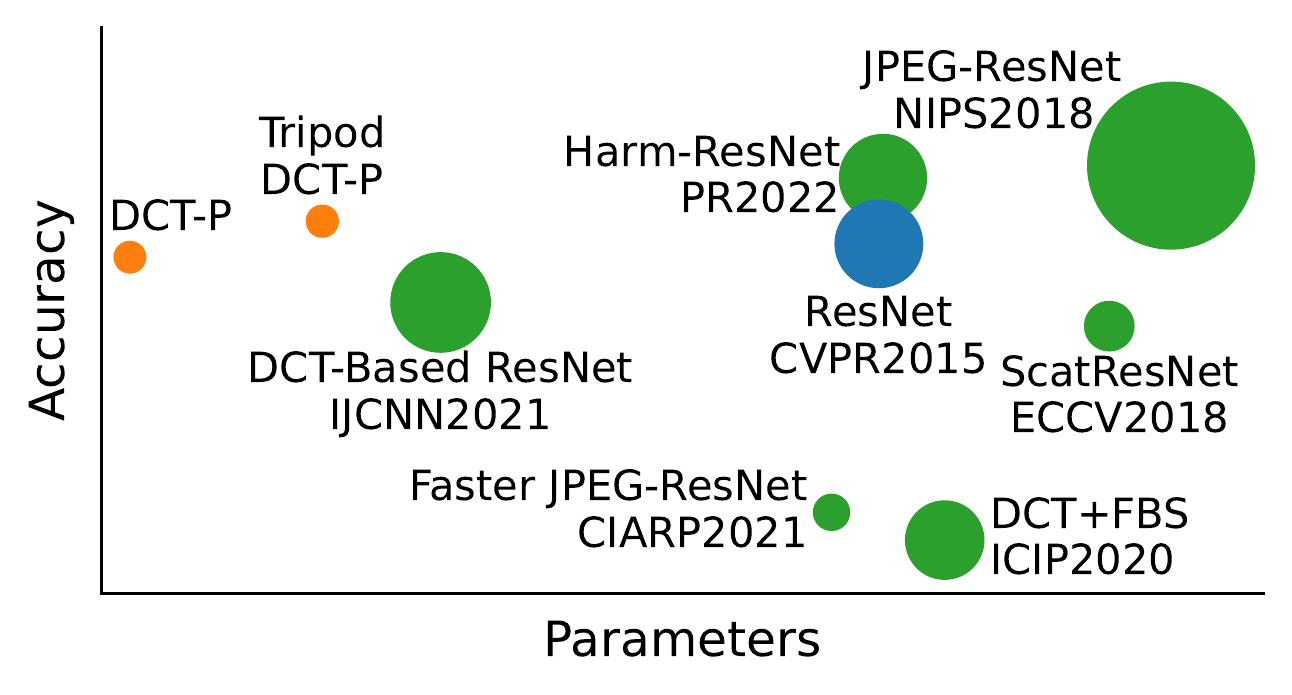}\vspace{-10pt}
    \caption{ImageNet-1K classification benchmark of the proposed methods (orange) versus some other transform-based methods (green). The size of a circle represents its computational cost. ResNet-50 (blue) is used as the backbone.\vspace{-5pt}}
    \label{fig: benchmark}
\end{figure}

\begin{figure}[htbp]
\centering
\subfloat[\label{fig: Conv Block}Conv blocks V1 and V2.]{\includegraphics[width=0.45\linewidth]{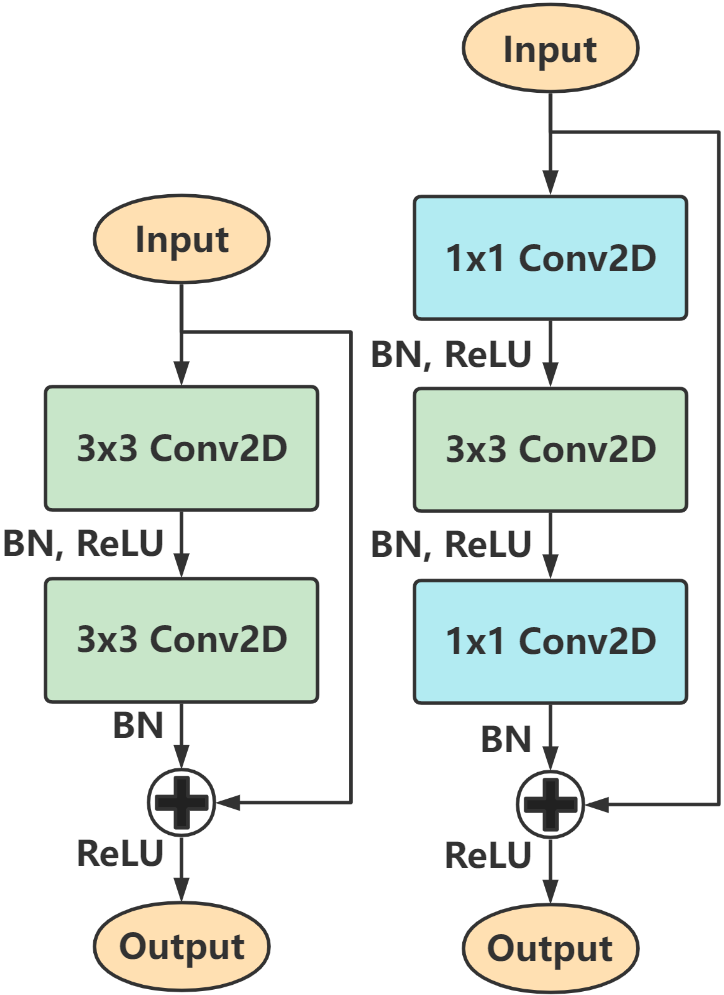}}\quad
\subfloat[\label{fig: DCT-P Residual Block} DCT-P blocks  V1 and V2.]{\includegraphics[width=0.45\linewidth]{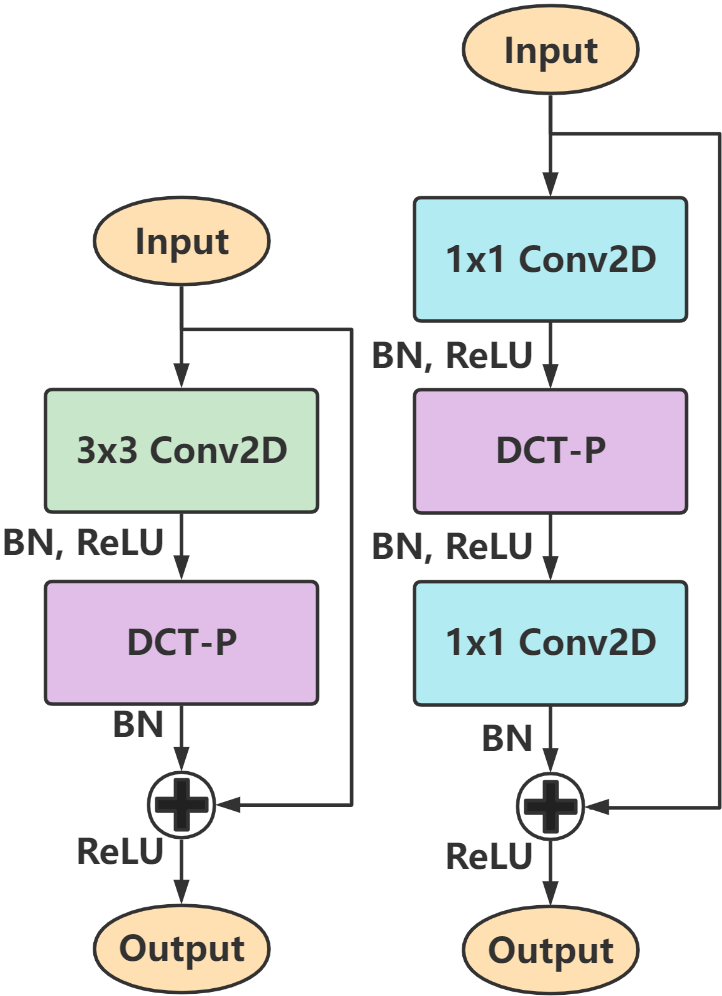}}\vspace{-5pt}
\caption{ResNet's convolutional residual blocks (a) versus the proposed DCT-Perceptron (DCT-P) residual blocks (b). BN stands for batch normalization. One $3\times3$ Conv2D layer in each block is replaced by one proposed DCT-perceptron layer. More than 21\% parameters and 32\% MACs are reduced for ResNet-50 with reaching comparable accuracy results on ImageNet-1K.}
\label{fig: block}
\end{figure}

Fourier convolution theorem states that the convolution $\mathbf{y}=\mathbf{x}*\mathbf{w}$ of two vectors {\bf x} and {\bf w} in the space domain can be implemented in the Discrete Fourier Transform (DFT) domain by elementwise multiplication: 
\begin{equation}
\mathbf{Y}(k) = \mathbf{X}(k)\cdot \mathbf{W}(k), \label{eq: DFT Convolution}
\end{equation}
where $\mathbf{Y}= \mathscr{F}(\mathbf{y}), \mathbf{X}= \mathscr{F}(\mathbf{x})$, and $\mathbf{W}= \mathscr{F}(\mathbf{w})$ are the DFTs of $\mathbf{y}, \mathbf{x}$, and $\mathbf{w}$, respectively, and $\mathscr{F}$ is the Fourier transform operator. Equation~\ref{eq: DFT Convolution} holds when the size of the DFT is larger than the size of the convolution output $\mathbf{y}$. In this paper, we use the Fourier convolution theorem to develop novel network layers. Since there is one-to-one relationship between the kernel coefficients $\mathbf{w}$ and the Fourier transform $\mathbf{W}$,     
kernel weights can be learned in the Fourier transform domain using backpropagation-type algorithms.

Since the DFT is a complex-valued transform, we replace the DFT with the real-valued Discrete Cosine Transform (DCT). Our DCT-based layer replaces the $3\times3$ convolutional layer of ResNet, as shown in Figure~\ref{fig: block}. 
DCT can express a vector in terms of a sum of cosine functions oscillating at different frequencies the DCT coefficients contain frequency domain information. 
DCT is the main enabler of a wide range of image and video coding standards including JPEG and MPEG \cite{wallace1991jpeg, le1991mpeg}. With our proposed layer, our models can reduce the number of parameters in ResNets significantly while producing comparable accuracy results as the original ResNets. This is possible because we can implement convolution-like operations in the DCT domain using only elementwise multiplications as in the Fourier transform. Both the DCT and the Inverse-DCT (IDCT) have fast $O(Nlog_2 N)$ algorithms which are actually faster than FFT because of the real-valued nature of the transform. An important property of the DCT is that convolutions can be implemented in the DCT domain using a modified version of the Fourier Convolution Theorem \cite{shen1998dct}.

% \begin{figure}[htbp]
% \centering
% \subfloat[\label{fig: Conv Block V1}Conv block V1.]{\includegraphics[width=0.28\linewidth]{figures/resnetv1.png}}\quad
% \subfloat[\label{fig: DCT-P Residual Block V1} DCT-P block V1.]{\includegraphics[width=0.29\linewidth]{figures/resnetdctv1.png}}\\
% \subfloat[\label{fig: Conv Block V2}Conv block V2.]{\includegraphics[width=0.28\linewidth]{figures/resnetv2.png}}\quad
% \subfloat[\label{fig: DCT-P Residual Block V2} DCT-P block V2.]{\includegraphics[width=0.285\linewidth]{figures/resnetdctv2.png}}
% \caption{ResNet's convolutional residual blocks (a) and (c) versus the proposed DCT-Perceptron (DCT-P) residual blocks (b) and (d), respectively. BN stands for batch normalization. We replace the $3\times3$ Conv2D layers with DCT-based layers whose input shape and output shape are the same as the proposed DCT-perceptron layers. }
% \label{fig: block}
% \end{figure}

\section{Related transform domain methods}

\noindent \textbf{FFT-based methods} The Fast Fourier transform (FFT) algorithm is the most important signal and image processing method. Convolutions in time and image domains can be performed using elementwise multiplications in the Fourier domain. However, the Discrete Fourier Transform (DFT) is a complex transform.  In~\cite{chi2020fast, mohammad2021substitution}, the fast Fourier Convolution (FFC) method is proposed in which the authors designed the FFC layer based on the so-called Real-valued Fast Fourier transform (RFFT). In RFFT-based methods, they concatenate the real and the imaginary parts of the FFT outputs, and they apply convolutions with ReLU in the concatenated Fourier domain. They do not take advantage of the Fourier domain convolution theorem.  Concatenating the real part and the imaginary part of the complex tensors increases the number of parameters and their model requires more parameters than the original ResNets because the number of channels is doubled after concatenating. Our method takes advantage of the convolution theorem and it can reduce the number of parameters significantly while producing comparable accuracy results.\\

\noindent \textbf{Wavelet-based methods} Wavelet transform (WT) is a well-established signal and image processing method and WT-based neural networks include~\cite{liu2019multi,oyallon2018compressing}. However, convolutions cannot be implemented in the wavelet domain.\\

\noindent \textbf{DCT-based methods} Other DCT-based methods include~\cite{gueguen2018faster, dos2020good, dos2021less, xu2021dct, ulicny2022harmonic}. Since the images are stored in the DCT domain in JPEG format, the authors use DCT coefficients of the JPEG images in~\cite{gueguen2018faster, dos2020good, dos2021less} but they did not use the transform domain convolution theorem to train the network. In~\cite{xu2021dct}, transform domain convolutional operations are used only during the testing phase. The authors did not train the kernels of the CNN in the DCT domain.  
%kernels are zero-padded to the same size as the input image, then DCT is applied on both padded kernels. DCT results of the input image and the padded kernels are element-wise multiplied to obtain the DCT of the output. Their kernels are not trained in the DCT domain.
They only take advantage of the fast DCT computation and reduce parameters by changing $3\times 3$ Conv2D layers to $2\times 2$. In our implementation, we train the filter parameters in the DCT domain and we use the soft-thresholding operator as the nonlinearity. It is not possible to train a network using DCT without soft-thresholding because both positive and negative amplitudes are important in the DCT domain. %\textcolor{blue}{In~\cite{chen2022discrete}, authors proposed a DCT-based pruning method that removes some convolutional kernels corresponding to the high-frequency part on the DCT feature map. The pruned network is still a CNN. It does not contain any DCT-based layer.} 
In~\cite{ulicny2022harmonic}, Harmonic convolutional networks based on DCT were proposed. Only forward DCT without inverse DCT computation is employed to obtain Harmonic blocks for feature extraction. In contrast with spatial convolution with learned kernels, this study proposes feature learning by weighted combinations of responses of predefined filters. The latter extracts harmonics from lower-level features in a region, and the latter layer applies DCT on the outputs from the previous layer which are already encoded by DCT.\\
    
% \noindent \textbf{HetConv blocks} Heterogeneous kernel-based convolutions (HetConv) method was first proposed by Pravendra \textit{et al.} in 2019 \cite{singh2019hetconv}. The HetConv uses many point-wise convolutions to replace $3\times 3$ convolutions. In 2020, Maneesh \textit{et al.} proposed a trimmed version of MobileNet architecture called Reduced MobileNet-V2~\cite{ayi2020rmnv2}. They replace bottleneck layers with HetConv blocks. Although HetConv reduced the parameters effectively, it is still based on regular convolution. % In this paper, we mainly compare our work with the frequency-analysis-based methods.\\
    
\noindent \textbf{Trainable soft-thresholding} The soft-thresholding function is widely used in wavelet transform domain denoising~\cite{donoho1995noising} and as a proximal operator for the $\ell_1$ norm~\cite{karakucs2020simulation}. With trainable threshold parameters, soft-thresholding and its variants can be employed as the nonlinear function in the frequency domain-based networks~\cite{badawi2021discrete,pan2021fast,pan2022block,pan2022deep}. In the frequency domain, ReLU and its variants are not good choices because they remove negative valued frequency components. This may cause issues when applying the inverse transform, as the negative valued components are as equivalently important as the positive components. On the contrary, the computational cost of the soft-thresholding is similar to the ReLU, and it kepdf both positive and negative valued frequency components exceeding a learnable threshold.

\section{Methodology}
In this section, we first review the discrete cosine transform (DCT). Then, we introduce how we design the DCT-perceptron layer. Finally, we compare the proposed DCT-perceptron layer with the conventional convolutional layer.
\subsection{Background: DCT Convolution Theorem}
Discrete cosine transform (DCT) is widely used for frequency analysis~\cite{ahmed1974discrete, strang1999discrete}. The type-II one-dimensional (1D) DCT and inverse DCT (IDCT), which are most generally used, are defined as follows:
\begin{equation}
    \mathbf{X}_k = \sum_{n=0}^{N-1}\mathbf{x}_n\text{cos}\left[\frac{\pi}{N}\left(n+\frac{1}{2}\right) k\right],\label{eq: DCT}
\end{equation}
\begin{equation}
    \mathbf{x}_n = \frac{1}{N}\mathbf{X}_0+\frac{2}{N}\sum_{k=1}^{N-1}\mathbf{X}_k\text{cos}\left[\frac{\pi}{N}\left(n+\frac{1}{2}\right)k\right],\label{eq: IDCT}
\end{equation}
for $0\leq k \leq N-1, 0\leq n \leq N-1,$ respectively. 

DCT and IDCT can be implemented using butterfly operations. For an $N$-length vector, the complexity of each 1D transform is $O(N\log_2 N)$. The 2D DCT is obtained from 1D DCT in a separable manner and a two-dimensional (2D) DCT can be implemented in a separable manner using 1D-DCTs for computational efficiency. The complexity of 2D-DCT and 2D-IDCT on an $N\times N$ image is $O(N^2\log_2 N)$~\cite{vetterli1985fast}.

Compared to the Discrete Fourier transform (DFT), DCT is a real-valued transform and it also decomposes a given signal or image according to its frequency content. Therefore, DCT is more suitable for deep neural networks in terms of computational cost. 

DFT convolutional theorem states that an input feature map $\mathbf{x}\in\mathbb{R}^N$ and a kernel $\mathbf{w}\in\mathbb{R}^K$ can be convolved as:
\begin{equation}
\mathbf{x} *_c\mathbf{w} = \mathscr{F}^{-1}\left(\mathscr{F}(\mathbf{x})\cdot\mathscr{F}(\mathbf{w})\right), 
\end{equation}
where, $\mathscr{F}$ stands for DFT and $\mathscr{F}^{-1}$ stands for IDFT. ``$*_c$" is the circular convolution operator and ``$\cdot$" is the element-wise multiplication. In practice, the circular convolution  ``$*_c$" can be converted to regular convolution ``*" by padding zeros to input signals properly and making the size of the DFT larger than the size of the convolution. Similarly, the DCT convolution theorem is 
\begin{equation}
\mathbf{x}*^s\mathbf{w} = \mathscr{D}^{-1}\left(\mathscr{D}(\mathbf{x})\cdot\mathscr{D}(\mathbf{w})\right), 
\end{equation}
where, $\mathscr{D}$ stands for DCT and $\mathscr{D}^{-1}$ stands for IDCT. ``$*^s$" stands for symmetric convolution and its relationship to the linear convolution is
\begin{equation}
    \mathbf{x}*^s\mathbf{w} = \mathbf{x}*\mathbf{\tilde{w}}
\end{equation}
where $\mathbf{\tilde{w}}$ is the symmetrically extended kernel in $\mathbb{R}^{2K-1}$:
\begin{equation}
    \mathbf{\tilde{w}}_k = \mathbf{w}_{|K-1-k|},
\end{equation}
for $k=0, 1, ..., 2K-2$. 
DCT and DFT convolution theorems can be extended to two or three dimensions in a straightforward manner. In the next subsection, we will describe how we will use DCT as a deep-neural network layer.
We call this new layer the DCT-Perceptron layer.

\subsection{DCT-Perceptron Layer}
\begin{figure}[htbp]
\centering
\subfloat[\label{fig: DCT-Px1}DCT-P.]{\includegraphics[width=0.2\linewidth]{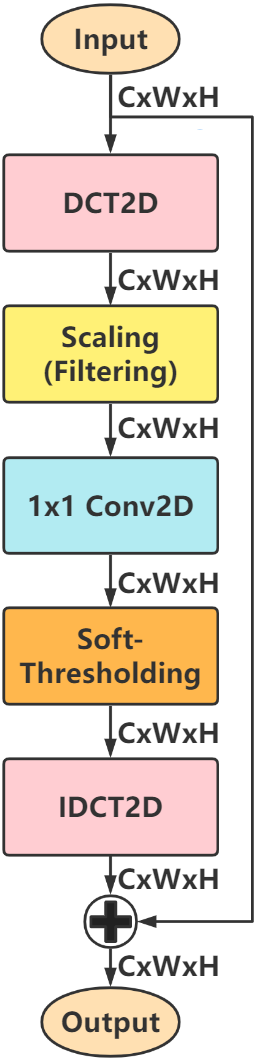}}\quad
\subfloat[\label{fig: DCT-Px3}Tripod DCT-P.]{\includegraphics[width=0.54\linewidth]{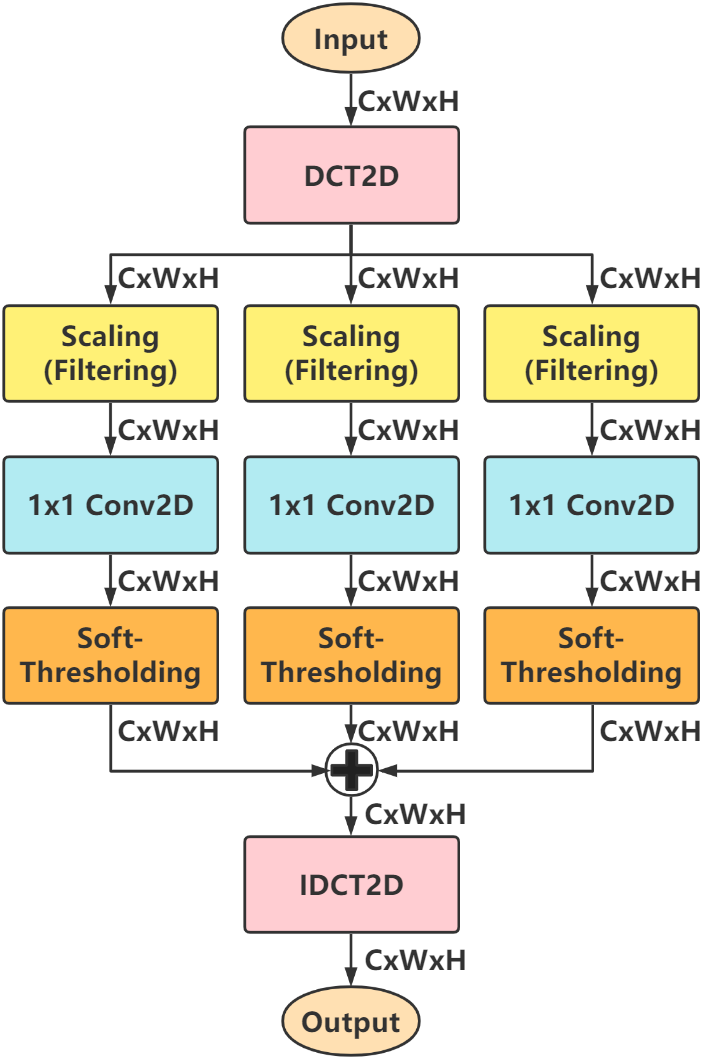}}\vspace{-5pt}
\caption{(a) DCT-perceptron  and (b) tripod DCT-perceptron. 
%Perceptron along the channel axis is equivalent to $1\times1$ Conv2D layer. 
Since the DCT2D and the IDCT2D are linear transforms, summation in the DCT domain and in the space domain are equivalent. For computational efficiency, the DCT2D and IDCT2D are implemented as two 1D transforms along the weight and the height, respectively. In our CIFAR-10 experiment the tripod structure reaches the highest accuracy among the multiple-pod structures.
}
\label{fig: DCT-P}
\end{figure}

An overview of the DCT-perceptron layer architecture is presented in Figure~\ref{fig: DCT-P}. We first compute $c$ 2D-DCTs of the input feature map tensor along the width ($W$) and height ($H$) axes. In the DCT domain, a scaling layer performs the filtering operation by element-wise multiplication. After this step, we have $1\times 1$ Conv2D layer, and a trainable soft-thresholding layer which denoises the data and acts as the nonlinearity of the DCT-perceptron as shown in Figure~\ref{fig: DCT-P}.

The scaling layer is derived from the property that the convolution in the space domain is equivalent to the elementwise multiplication in the transform domain. In detail, given an input tensor in $\mathbb{R}^{C\times W \times H}$, we element-wise multiply it with a weight matrix in $\mathbb{R}^{W \times H}$ to perform an operation equivalent to the spatial domain convolutional filtering. 

The trainable soft-thresholding layer is applied to remove small entries in the DCT domain. It is similar to image coding and transform domain denoising. It is defined as:
\begin{equation}
    \mathbf{y} = \mathcal{S}_T(\mathbf{x}) = \text{sign}(\mathbf{x})(|\mathbf{x}|-T)_+,
\end{equation}
where, $T$ is a non-negative trainable threshold parameter. For an input tensor in $\mathbb{R}^{C\times W \times H}$, there are $W\times H$ different threshold parameters. The threshold parameters are determined using the back-propagation algorithm. Our CIFAR-10 experiments show that the soft-thresholding is superior to the ReLU as the non-linear function in DCT analysis. 

Overall, the DCT-perceptron layer is computed as:
\begin{equation}
    \mathbf{y}=\mathscr{D}^{-1}\left(\mathcal{S}_\mathbf{T}\left(\mathscr{D}(\mathbf{x})\cdot\mathbf{V})\otimes\mathbf{H}\right)\right)+\mathbf{x},
\end{equation}
%and the $P$-pod DCT-perceptron layer is computed as:
%\begin{equation}
%    \mathbf{y}=\mathscr{D}^{-1}\left(\sum_{i=0}^{P-1}\left(\mathcal{S}_{\mathbf{T}_i}\left(\mathscr{D}(\mathbf{x})\cdot\mathbf{v}_i*\mathbf{k}_i\right)\right)\right),
%\end{equation}
where, $\mathscr{D}$ and $\mathscr{D}^{-1}$ stand for 2D-DCT and 2D-IDCT, $\mathbf{V}$ is the scaling matrix, $\mathbf{H}$ are the kernels in the $1\times1$ Conv2D layer, $\mathbf{T}$ is the threshold parameter matrix in the soft-thresholding, ``$\cdot$" stands for the element-wise multiplication, and $\otimes$ represents a $1\times 1$ 2D convolution over an input composed of several input planes performed using PyTorch's Conv2D API. 
%``$*$" is same as the dot-product because the size of the convolution kernels is $1\times1$. 

\begin{figure}[t]
\centering
\includegraphics[width=0.8\linewidth]{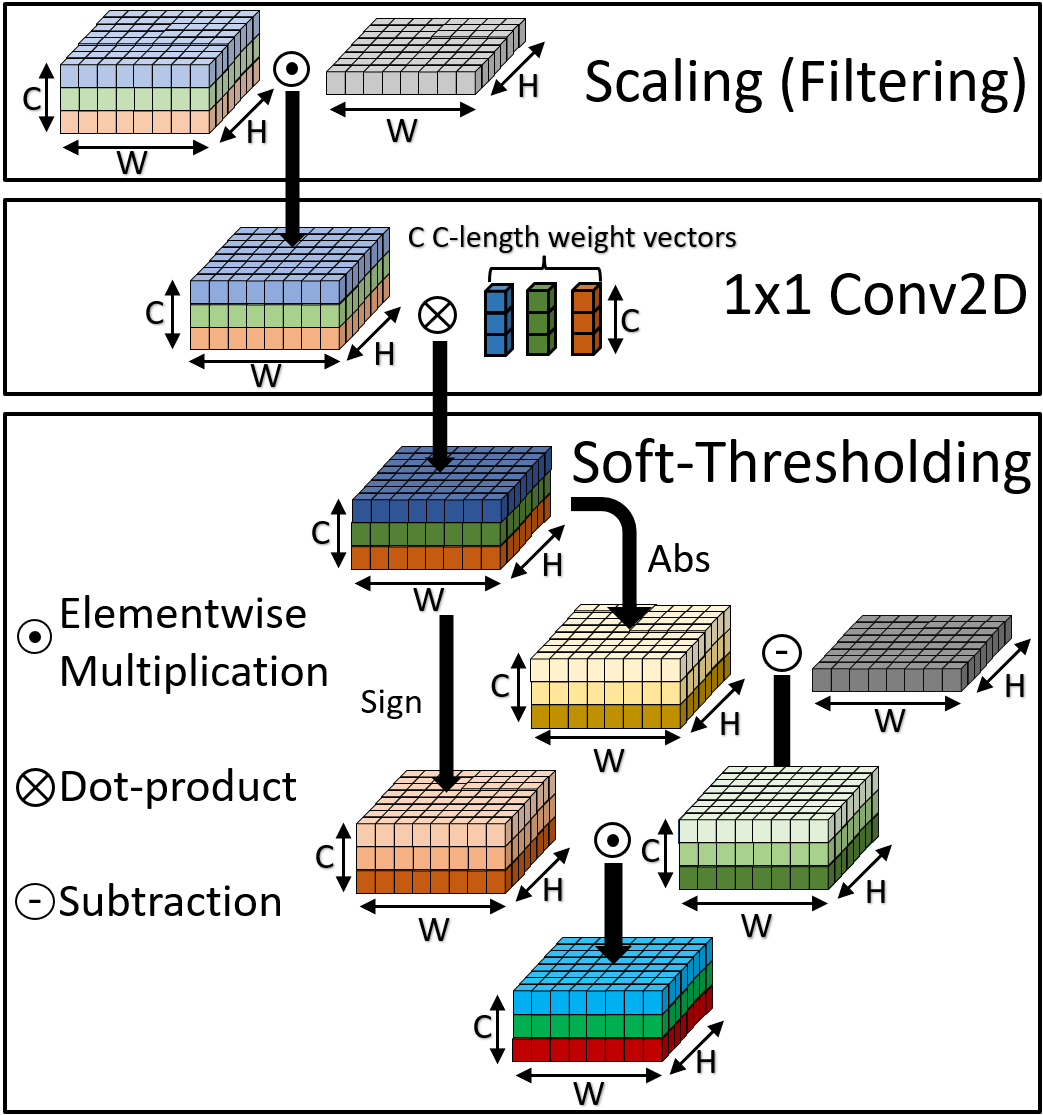}\vspace{-5pt}
\caption{Processing in the DCT domain. Multiplications in the soft-thresholding block are efficiently implemented using sign-bit operations.\vspace{-5pt}}
\label{fig: pod}
\end{figure}

We further extend the DCT-perceptron layer to a tripod structure to get higher accuracy with more trainable parameters (still significantly fewer than Conv2D) as shown in Figure~\ref{fig: DCT-Px3}. The process of layers for the DCT domain analysis (scaling, $1\times1$ Conv2D, and soft-thresholding) is depicted in Figure~\ref{fig: pod}.

In this section, we compare the difference between the DCT-perceptron layer and the Conv2D layer of ResNet. To compare the computational cost and the number of parameters, we specificity stride to 1 and apply padding=``SAME" on the Conv2D layers.
%\subsubsection{Characteristics Comparison}

%As Figure~\ref{fig: 2D-convolution VS 2D-DCT} shows, in a Conv2D layer, each output is computed using a limited region of the input tensor slide in each time. On the other hand, in a DCT-perceptron layer, each output contains the information from the entire tensor slide, because each entry is a discrete-cosine-weighted summation on the entire tensor. Therefore, parameters in the DCT-preceptron layer are trained using whole information, while the convolutional parameters in the Conv2D are updated using using limited information, in each time.

\begin{figure}[htbp]
\centering
\subfloat[Conv2D]{\includegraphics[width=0.4\linewidth]{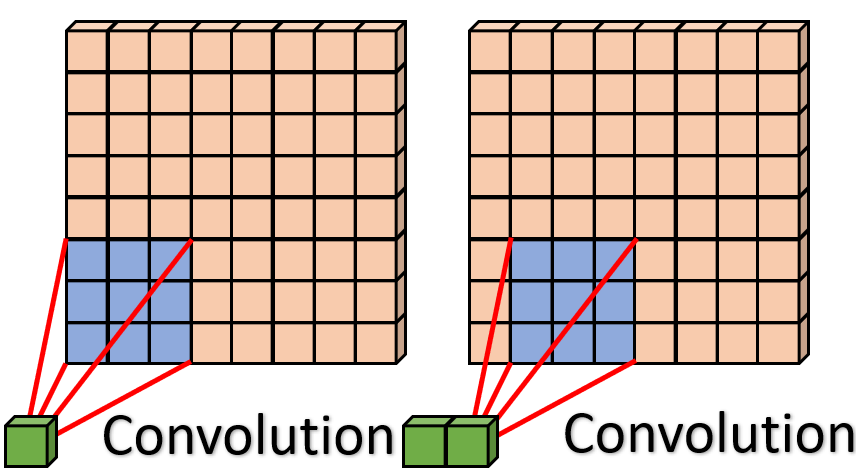}}\quad
\subfloat[DCT-Perceptron]{\includegraphics[width=0.4\linewidth]{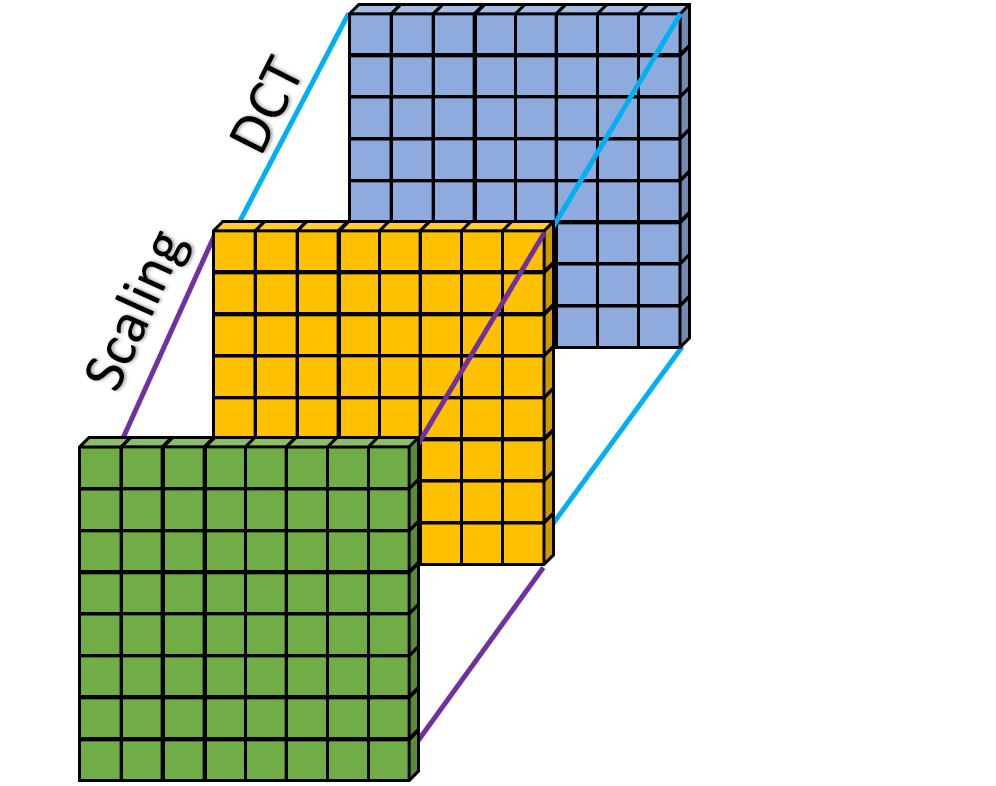}}\vspace{-5pt}
\caption{To compute different entries of an output slice, the Conv2D layer uses same weights while the DCT-perceptron layer uses different. Thus, the Conv2D layer is spatial-agnostic while the DCT-Perceptron layer is location-specific.\vspace{-5pt}}
\label{fig: 2D-convolution VS 2D-DCT}
\end{figure}

As Figure~\ref{fig: 2D-convolution VS 2D-DCT} shows,  convolution has spatial-agnostic and channel-specific characteristics in ResNet's Conv2D layer. Because of the spatial-agnostic characteristics of a Conv2D layer, the network cannot adapt different visual patterns corresponding to different spatial locations. 
%as the same kernel parameters are multiplied by different entries on different width and height on the spatial feature map. 
On the contrary, the DCT-perceptron layer is location-specific and channel-specific. 2D-DCT is location-specific but channel-agnostic, as the 2D-DCT is computed using the entire block as a weighted summation on the spatial feature map. The scaling layer is also location-specific but channel-agnostic, as different scaling parameters (filters) are applied on different entries, and weights are shared in different channels. That is why we also use Pytorch's $1\times 1$ Conv2D  to make the DCT-perceptron layer channel-specific.

\begin{table*}[htbp]
    \centering
    \begin{tabular}{lcc}
    \hline\noalign{\smallskip}
         \bf{Layer (Operation)}&\bf{Parameters}&\bf{MACs}\\
         \noalign{\smallskip}\hline\noalign{\smallskip}
         $K\times K$ Conv2D& $K^2C^2$&$K^2N^2C^2$\\
         $3\times 3$ Conv2D& $9C^2$&$9N^2C^2$\\
         \hline
         DCT2D&$0$& $(\frac{5N^2}{2}\log_2 N+\frac{N^2}{3}-6N+\frac{62}{3})C$\\
         
         Scaling, Soft-Thresholding& $2N^2$&$N^2C$\\
         $1\times 1$ Conv2D & $C^2$ & $N^2C^2$\\
         
         IDCT2D& $0$& $(\frac{5N^2}{2}\log_2 N+\frac{N^2}{3}-6N+\frac{62}{3})C$\\
         
         \bf{DCT-Perceptron} & $2N^2+C^2$&$(5N^2\log_2 N+\frac{5N^2}{3}-6N+\frac{124}{3})C+N^2C^2$\\
         
         \bf{Tripod DCT-Perceptron} & $6N^2+3C^2$&$(5N^2\log_2 N+\frac{11N^2}{3}-6N+\frac{124}{3})C+3N^2C^2$\\
         \noalign{\smallskip}\hline\noalign{\smallskip}
    \end{tabular}\vspace{-10pt}
    \caption{Parameters and Multiply–Accumulate (MACs) for a $C$-channel $N\times N$ image.\vspace{-10pt}} 
    \label{tab: parameters and MACs}
\end{table*}

% \noindent \textbf{Computational Cost Comparison}
The comparison of the number of parameters and  Multiply–Accumulate (MACs) is presented in Table~\ref{tab: parameters and MACs}. In a Conv2D layer, we need to compute $N^2K^2$ multiplications for a $C$-channel feature map of size $N\times N$ with a kernel size of $K\times K$. In the DCT-perceptron layer, there are $N^2$ multiplications from the scaling and $N^2$ multiplications from the $1\times1$ Conv2D layer. There is no multiplication in the soft-thresholding because the product between the sign of the input and the subtraction result can be implemented using the sign-bit operations only.  There are $\frac{N^2}{2}\log_2 N+\frac{N^2}{3}-2N+\frac{8}{3}$ multiplications with $\frac{5N^2}{2}\log_2 N+\frac{N^2}{3}-6N+\frac{62}{3}$ additions to implement a 2D-DCT or a 2D-IDCT using the fast DCT algorithms with butterfly operations~\cite{vetterli1985fast}. Therefore, there are totally $(5N^2\log_2 N+\frac{5N^2}{3}-6N+\frac{124}{3})C+N^2C^2$ MACs in a single-pod DCT-perceptron layer and $(5N^2\log_2 N+\frac{11N^2}{3}-6N+\frac{124}{3})C+3N^2C^2$ MACs in a tripod DCT-perceptron layer. Compared to the ResNet's Conv2D layer with $9N^2C^2$ MACs, the DCT-perceptron layer reduces $6N^2C^2$ MACs to $(5N^2\log_2 N+\frac{11N^2}{3}-6N+\frac{124}{3})C$ MACs. In other words, it reduces $O(N^2C^2)$ to $O((N^2\log_2N)C)$ if we omit $3N^2C^2$ MACs. This significantly reduces the computation cost because $C$ is much larger than $N$ in most of the hidden layers of ResNets. A tripod DCT-perceptron layer only has $2N^2C+2N^2C^2$ more MACs than a single-pod DCT-perceptron layer because the 2D-DCT in each pod is equivalent to each other.

\subsection{Introducing DCT-Perceptron to ResNets}
ResNet-20 and ResNet-18 use convolutional residual block V1 in Fig.~\ref{fig: block}, while ResNet-50 uses block V2 as shown in Fig.~\ref{fig: block}. We replace some $3\times 3$ Conv2D layers whose input and output shapes are the same in the ResNets' convolutional residual blocks by the proposed DCT-perceptron layer. In ResNet-20 and ResNet-18, we replace the second $3\times 3$ Conv2D layer in each convolution block. In ResNet-50, we replace $3\times 3$ Conv2D layers in convolution blocks except for Conv2\_1, Conv3\_1, Conv4\_1, and Conv5\_1, as the downsampling is performed by them. We call those revised using the DCT-perceptron layer as DCT-ResNets and those revised using the tripod DCT-perceptron layer as Tripod-DCT-ResNets. Details of the DCT-ResNets are presented in Tables~\ref{tab: resnet-20},~\ref{tab: resnet-18} and~\ref{tab: resnet-50}, respectively. In our ablation study on CIFAR-10, we also replace more $3\times 3$ Conv2D layers, but the experiments show that replacing more layers leads to drops in accuracy as these models have fewer parameters. 

% \begin{table}[htbp]
% \centering
%     \begin{tabular}{ccc}
%     \hline\noalign{\smallskip}
% 		\bf{Layer}&\bf{Output Shape}&\bf{Implementation Details}\\
%          \noalign{\smallskip}\hline\noalign{\smallskip}
% 		Input&$3\times32\times32$&-\\
% 		Conv1&$16\times32\times32$&$3\times3, 16$\\
% 		Conv2\_x&$16\times32\times32$&$\left[ \begin{array}{c} 3\times3, 16  \\ \bf{3\times3, 16} \end{array}\right]\times 3$\\
% 		Conv3\_x&$32\times16\times16$&$\left[ \begin{array}{c} 3\times3, 32  \\ \bf{3\times3, 32} \end{array}\right]\times 3$\\
% 		Conv4\_x&$64\times8\times8$&$\left[ \begin{array}{c} 3\times3, 32  \\ \bf{3\times3, 64} \end{array}\right]\times 3$\\
% 		GAP&$64$&Global Average Pooling\\
% 		Output&$10$&Dense (unit = 10)\\
%     \noalign{\smallskip}\hline\noalign{\smallskip}
% 	\end{tabular}
% 	\caption{Structure of ResNet-20 for CIFAR-10. Building blocks are shown in brackets, with the numbers of blocks stacked. Downsampling is performed by Conv3\_1 and Conv4\_1 with a stride of 2. Bold layers are replaced by the proposed DCT-perceptron layer.}
% \label{tab: resnet-20}
% \end{table}

\begin{table}[htbp]
\centering
    \begin{tabular}{lcc}
    \hline\noalign{\smallskip}
		\bf{Layer}&\bf{Output Shape}&\bf{Implementation Details}\\
         \noalign{\smallskip}\hline\noalign{\smallskip}
		%Input&$3\times32\times32$&-\\
		Conv1&$16\times32\times32$&$3\times3, 16$\\
		Conv2\_x&$16\times32\times32$&$\left[ \begin{array}{c} 3\times3, 16  \\ \bf{\text{DCT-P}, 16} \end{array}\right]\times 3$\\
		Conv3\_x&$32\times16\times16$&$\left[ \begin{array}{c} 3\times3, 32  \\ \bf{\text{DCT-P}, 32} \end{array}\right]\times 3$\\
		Conv4\_x&$64\times8\times8$&$\left[ \begin{array}{c} 3\times3, 32  \\ \bf{\text{DCT-P}, 64} \end{array}\right]\times 3$\\
		GAP&$64$&Global Average Pooling\\
		Output&$10$&Linear\\
    \noalign{\smallskip}\hline\noalign{\smallskip}
	\end{tabular}\vspace{-10pt}
	\caption{Structure of DCT-ResNet-20 for the CIFAR-10 classification task. Building blocks are shown in brackets, with the numbers of blocks stacked. Downsampling is performed by Conv3\_1 and Conv4\_1 with a stride of 2. We name layers following Table 1 in\cite{he2016deep} for the comparison of structures easily.}
\label{tab: resnet-20}
\end{table}

% \begin{table}[htbp]
%     \centering
%     \begin{tabular}{ccc}
%     \hline\noalign{\smallskip}
% 		\bf{Layer}&\bf{Output Shape}&\bf{Implementation Details}\\
%         \noalign{\smallskip}\hline\noalign{\smallskip}
% 		Input&$3\times224\times224$&-\\
% 		Conv1&$64\times112\times112$&$7\times7, 64$, stride 2\\
% 		MaxPool&$64\times56\times56$&$2\times2$, stride 2\\
% 		Conv2\_x&$64\times56\times56$&$\left[ \begin{array}{c} 3\times3, 64  \\ \bf{3\times3, 64} \end{array}\right]\times 2$\\
% 		Conv3\_x&$128\times28\times28$&$\left[ \begin{array}{c} 3\times3, 128  \\ \bf{3\times3, 128} \end{array}\right]\times 2$\\
% 		Conv4\_x&$256\times14\times14$&$\left[ \begin{array}{c} 3\times3, 256  \\ \bf{3\times3, 256} \end{array}\right]\times 2$\\
% 		Conv5\_x&$512\times7\times7$&$\left[ \begin{array}{c} 3\times3, 512  \\ \bf{3\times3, 512} \end{array}\right]\times 2$\\
% 		GAP&$512$&Global Average Pooling\\
% 		Output&$1000$&Dense (unit = 1000)\\
%     \noalign{\smallskip}\hline\noalign{\smallskip}
% 	\end{tabular}
% 	\caption{Structure of ResNet-18 for ImageNet-1K. Building blocks are shown in brackets, with the numbers of blocks stacked. Downsampling is performed by Conv3\_1, Conv4\_1 and Conv5\_1 with a stride of 2. Bold layers are replaced by the proposed DCT-perceptron layer.}
% \label{tab: resnet-18}
% \end{table}

\begin{table}[htbp]
    \centering
    \begin{tabular}{lcc}
    \hline\noalign{\smallskip}
		\bf{Layer}&\bf{Output Shape}&\bf{Implementation Details}\\
        \noalign{\smallskip}\hline\noalign{\smallskip}
		%Input&$3\times224\times224$&-\\
		Conv1&$64\times112\times112$&$7\times7, 64$, stride 2\\
		MaxPool&$64\times56\times56$&$2\times2$, stride 2\\
		Conv2\_x&$64\times56\times56$&$\left[ \begin{array}{c} 3\times3, 64  \\ \bf{\text{DCT-P}, 64} \end{array}\right]\times 2$\\
		Conv3\_x&$128\times28\times28$&$\left[ \begin{array}{c} 3\times3, 128  \\ \bf{\text{DCT-P}, 128} \end{array}\right]\times 2$\\
		Conv4\_x&$256\times14\times14$&$\left[ \begin{array}{c} 3\times3, 256  \\ \bf{\text{DCT-P}, 256} \end{array}\right]\times 2$\\
		Conv5\_x&$512\times7\times7$&$\left[ \begin{array}{c} 3\times3, 512  \\ \bf{\text{DCT-P}, 512} \end{array}\right]\times 2$\\
		GAP&$512$&Global Average Pooling\\
		Output&$1000$&Linear\\
    \noalign{\smallskip}\hline\noalign{\smallskip}
	\end{tabular}\vspace{-10pt}
	\caption{Structure of DCT-ResNet-18 for the ImageNet-1K classification task. Building blocks are shown in brackets, with the numbers of blocks stacked. Downsampling is performed by Conv3\_1, Conv4\_1 and Conv5\_1 with a stride of 2.\vspace{-10pt}}% We name layers same as Table 1 in\cite{he2016deep} for the comparison of structures easily.}
\label{tab: resnet-18}
\end{table}

% \begin{table}[htbp]
%     \centering
%     \begin{tabular}{ccc}
%     \hline\noalign{\smallskip}
% 		\bf{Layer}&\bf{Output Shape}&\bf{Implementation Details}\\
%         \noalign{\smallskip}\hline\noalign{\smallskip}
% 		Input&$3\times224\times224$&-\\
% 		Conv1&$64\times112\times112$&$7\times7, 64$, stride 2\\
% 		MaxPool&$64\times56\times56$&$2\times2$, stride 2\\
% 		Conv2\_x&$256\times56\times56$&$\left[ \begin{array}{c} 1\times1, 64  \\ \bf{3\times3, 64} \\ 1\times1, 256 \end{array}\right]\times 3$\\
% 		Conv3\_x&$512\times28\times28$&$\left[ \begin{array}{c} 1\times1, 128  \\ \bf{3\times3, 128} \\ 1\times1, 512 \end{array}\right]\times 4$\\
% 		Conv4\_x&$1024\times14\times14$&$\left[ \begin{array}{c} 1\times1, 256  \\ \bf{3\times3, 256} \\ 1\times1, 1024 \end{array}\right]\times 6$\\
% 		Conv5\_x&$2048\times7\times7$&$\left[ \begin{array}{c} 1\times1, 512  \\ \bf{3\times3, 512} \\ 1\times1, 2048 \end{array}\right]\times 3$\\		
% 		GAP&$2048$&Global Average Pooling\\
% 		Output&$1000$&Dense (unit = 1000)\\
%     \noalign{\smallskip}\hline\noalign{\smallskip}
% 	\end{tabular}
% 	\caption{Structure of ResNet-50 for ImageNet-1K. Building blocks are shown in brackets, with the numbers of blocks stacked. Downsampling is performed by Conv3\_1, Conv4\_1 and Conv5\_1 with a stride of 2. These layers are retained because their input and output size are different. The remaining convolutional layers in bold are replaced by the proposed DCT-perceptron layer.}
% \label{tab: resnet-50}
% 	\end{table}

\begin{table}[htbp]
    \centering
    \begin{tabular}{lcc}
    \hline\noalign{\smallskip}
		\bf{Layer}&\bf{Output Shape}&\bf{Implementation Details}\\
        \noalign{\smallskip}\hline\noalign{\smallskip}
		%Input&$3\times224\times224$&-\\
		Conv1&$64\times112\times112$&$7\times7, 64$, stride 2\\
		MaxPool&$64\times56\times56$&$2\times2$, stride 2\\
		Conv2\_1&$256\times56\times56$&$\left[ \begin{array}{c} 1\times1, 64  \\ 3\times3, 64, \text{stride }2 \\ 1\times1, 256 \end{array}\right]$\\
		Conv2\_x&$256\times56\times56$&$\left[ \begin{array}{c} 1\times1, 64  \\ \bf{\text{DCT-P}, 64} \\ 1\times1, 256 \end{array}\right]\times 2$\\
		Conv3\_1&$512\times28\times28$&$\left[ \begin{array}{c} 1\times1, 128  \\ 3\times3, 128, \text{stride }2 \\ 1\times1, 512 \end{array}\right]$\\
		Conv3\_x&$512\times28\times28$&$\left[ \begin{array}{c} 1\times1, 128  \\ \bf{\text{DCT-P}, 128} \\ 1\times1, 512 \end{array}\right]\times 3$\\
		Conv4\_1&$1024\times14\times14$&$\left[ \begin{array}{c} 1\times1, 256  \\ 3\times3, 256, \text{stride }2 \\ 1\times1, 1024 \end{array}\right]$\\
		Conv4\_x&$1024\times14\times14$&$\left[ \begin{array}{c} 1\times1, 256  \\ \bf{\text{DCT-P}, 256} \\ 1\times1, 1024 \end{array}\right]\times 5$\\
		Conv5\_1&$2048\times7\times7$&$\left[ \begin{array}{c} 1\times1, 512  \\ 3\times3, 512, \text{stride }2 \\ 1\times1, 2048 \end{array}\right]$\\		
		Conv5\_x&$2048\times7\times7$&$\left[ \begin{array}{c} 1\times1, 512  \\ \bf{\text{DCT-P}, 512} \\ 1\times1, 2048 \end{array}\right]\times 2$\\		
		GAP&$2048$&Global Average Pooling\\
		Output&$1000$&Linear\\
    \noalign{\smallskip}\hline\noalign{\smallskip}
	\end{tabular}\vspace{-10pt}
	\caption{Structure of DCT-ResNet-50 for the ImageNet-1K classification task. Building blocks are shown in brackets, with the numbers of blocks stacked.\vspace{-10pt}} %We name layers the same as Table 1 in\cite{he2016deep}.}
\label{tab: resnet-50}
	\end{table}
	
\section{Experimental Results}
Our experiments are carried out on a workstation computer with an NVIDIA RTX 3090 GPU. The code is written in PyTorch in Python 3. First, we experiment on the CIFAR-10 dataset with ablation studies. Then, we compare DCT-ResNet-18 and DCT-ResNet-50 with other related works on the ImageNet-1K dataset. We also insert an additional DCT-perceptron layer with batch normalization before the global average pooling layer. The additional DCT-perceptron layer improves the accuracy of  the original ResNets.

\begin{table*}[htbp]
    \centering
    \begin{tabular}{lcc}
    \hline\noalign{\smallskip}
       \bf{Method}&\bf{Parameters}&\bf{Accuracy}\\
        \noalign{\smallskip}\hline\noalign{\smallskip}
        ResNet-20 \cite{he2016deep} (official)&0.27M&91.25\%\\
        ResNet-20 (our trial, baseline)&272,474&91.66\%\\
        \bf{DCT-ResNet-20}&\bf{151,514 (44.39\%$\downarrow$)}&\bf{91.59\%}\\
        DCT-ResNet-20 (without shortcut connection in DCT-Perceptron)&151,514 (44.39\%$\downarrow$)&91.12\%\\
        DCT-ResNet-20 (without scaling in DCT-Perceptron)&147,482 (45.87\%$\downarrow$)&90.46\%\\
        DCT-ResNet-20 (using ReLU with thresholds in DCT-Perceptron)&151,514 (44.39\%$\downarrow$)&91.30\%\\
        DCT-ResNet-20 (using ReLU in DCT-Perceptron)&147,818 (45.75\%$\downarrow$)&91.06\%\\
        DCT-ResNet-20 (replacing all Conv2D)&51,034 (81.31\%$\downarrow$)&85.74\%\\
        Bipod-DCT-ResNet-20&175,706 (35.51\%$\downarrow$)&91.42\%\\
        \bf{TriPod-DCT-ResNet-20}&\bf{199,898 (26.64\%$\downarrow$)}&\bf{91.75\%}\\
        Tripod-DCT-ResNet-20 (with shortcut connection in tripod DCT-Perceptron)&199,898 (26.64\%$\downarrow$)&91.50\%\\
        % Tripod-DCT-ResNet-20 (replacing all Conv2D)&148,442 (45.52\%$\downarrow$)&86.70\%\\
        % Tripod-DCT-ResNet-20 (replacing all more Conv2D)&156,122 (42.70\%$\downarrow$)&89.94\%\\
        Quadpod-DCT-ResNet-20&224,090 (17.76\%$\downarrow$)&91.48\%\\
        Quintpod-DCT-ResNet-20&248,282 (8.88\%$\downarrow$)&91.47\%\\
       % DCT-ResNet-20x6&272,474 (0.00\%=)&91.47\%\\
        \bf{ResNet-20+1DCT-P}&\bf{276,826 (1.60\%$\uparrow$)}&\bf{91.82\%}\\
        
        \noalign{\smallskip}\hline\noalign{\smallskip}
    \end{tabular}\vspace{-10pt}
    \caption{Ablation study: CIFAR-10 classification Results.}
    \label{tab: CIFAR-10}
\end{table*}

\begin{table*}[htbp]
    \centering
    \begin{tabular}{lcccc}
    \hline\noalign{\smallskip}
        \bf{Method}&\bf{Parameters (M)}&\bf{MACs (G)}&\bf{Center-Crop Top1 Acc.}&\bf{Top5 Acc.}\\
        \noalign{\smallskip}\hline\noalign{\smallskip}
        ResNet-18 \cite{he2016deep} (Torchvision~\cite{torchvision}, baseline)&11.69&1.822&69.76\%&89.08\%\\
        %ResNet-18 baseline in~\cite{xu2021dct}&25.6M&69.34\%&89.02%\\
        DCT-based ResNet-18~\cite{xu2021dct}&8.74&-&68.31\%&88.22\%\\
         %ResNet-18 (our trial, baseline)&11,689,512&69.99\%&89.45\%\\
         \bf{DCT-ResNet-18}&\bf{6.14 (47.5\%$\downarrow$)}&\bf{1.374 (24.6\%$\downarrow$)}&\bf{67.84\%}&\bf{87.73\%}\\
         \bf{Tripod-DCT-ResNet-18}&\bf{7.56 (35.3\%$\downarrow$)}&\bf{1.377 (24.4\%$\downarrow$)}&\bf{69.55\%}&\bf{89.04\%}\\
         \bf{Tripod-DCT-ResNet-18+1DCT-P}&\bf{7.82 (33.1\%$\downarrow$)}&\bf{1.378 (24.4\%$\downarrow$)}&\bf{69.88\%}&\bf{89.00\%}\\
        \hline
        ResNet-50 \cite{he2016deep} (official~\cite{DeepResidualNetworksOfficialCode})&25.56&3.86&75.3\%&92.2\%\\
        ResNet-50 \cite{he2016deep} (Torchvision~\cite{torchvision})&25.56&4.122&76.13\%&92.86\%\\
        Order 1 + ScatResNet-50~\cite{oyallon2018compressing}&27.8&-&74.5\%&92.0\%\\
        %ResNet-50 baseline in~\cite{gueguen2018faster}&25.6M&75.78\%&92.65\%\\
        JPEG-ResNet-50~\cite{gueguen2018faster} &28.4&5.4&76.06\%&93.02\%\\
        ResNet-50+DCT+FBS (3x32)~\cite{dos2020good} &26.2&3.68&70.22\%&-\\
        ResNet-50+DCT+FBS (3x16)~\cite{dos2020good} &25.6&3.18&67.03\%&-\\
        Faster JPEG-ResNet-50~\cite{dos2021less} &25.1&2.86&70.49\%&-\\
        %ResNet-50 baseline in~\cite{xu2021dct}&25.6M&75.88\%&92.85\\
        DCT-based ResNet-50~\cite{xu2021dct}&21.30&-&74.73\%&92.30\%\\
       % ResNet-50 baseline in~\cite{ulicny2022harmonic}&25.6M&75.66\%&92.70\%\\
        Harm-ResNet-50~\cite{ulicny2022harmonic}&25.6&-&75.94\%&92.88\%\\
        %Harm-ResNet-50, progr.$\lambda$~\cite{ulicny2022harmonic}&19.7&-&75.38\%&92.56\%\\
        %ResNet-50 baseline in~\cite{chi2020fast}&25.6M&76.3\%&-\\
        FFC-ResNet-50 (+LFU, $\alpha=0.25$)~\cite{chi2020fast}&26.7&4.3&77.8\%&-\\
        FFC-ResNet-50 ($\alpha=1$)~\cite{chi2020fast}&34.2&5.6&75.2\%&-\\
        %ResNet-50 with DCT-pruning~\cite{chen2022discrete}&15.79&2.41&75.12\%&92.15\%\\
        
        ResNet-50 (our trial, baseline)&25.56&4.122&76.06\%&92.85\%\\
        \bf{DCT-ResNet-50}&\bf{18.28 (28.5\%$\downarrow$)}&\bf{2.772 (32.8\%$\downarrow$)}&\bf{75.17\%}&\bf{92.47\%}\\
        \bf{Tripod-DCT-ResNet-50}&\bf{20.15 (21.1\%$\downarrow$)}&\bf{2.780 (32.6\%$\downarrow$)}&\bf{75.52\%}&\bf{92.56\%}\\
        \bf{Tripod-DCT-ResNet-50+1DCT-P}&\bf{24.35 (4.7\%$\downarrow$)}&\bf{2.783 (32.5\%$\downarrow$)}&\bf{75.82\%}&\bf{92.76\%}\\
        \noalign{\smallskip}\hline\noalign{\smallskip}
    \end{tabular}\vspace{-10pt}
    \caption{ImageNet-1K center-crop results. We use Torchvision's official ResNet-18 model as the ResNet-18 baseline since we use PyTorch official training code with the default setting to train revised ResNet-18s. Other frequency-analysis-based methods are listed for comparison. %In Harm-ResNet-50~\cite{ulicny2022harmonic}, progr.$\lambda$ means the network applies progressive selection that omits a higher number of frequencies (determined by $\lambda$) in deeper layers to reduce parameters. 
    In FFC-ResNet-50 (+LFU, $\alpha=0.25$)~\cite{chi2020fast}, they use their proposed Fourier Units (FU) and local Fourier units (LFU) to process only 25\% channels of each input tensor and use regular $3\times 3$ Conv2D to process the remaining to get the accuracy of 77.8\% but top 5 accuracy result was not reported. 
    %This is quite different from our revision, as we use the proposed DCT-perceptron layer to process all channels. 
    We can also use a combination of regular $3\times 3$ Conv2D layers and DCT-perceptron layers to increase the accuracy by increasing the parameters. 
    According to the ablation study in~\cite{chi2020fast}, if all channels are processed by their proposed FU, FFC-ResNet-50 ($\alpha=1$) only gets the accuracy of 75.2\%. The reported number of parameters and MACs are much higher than the baseline in~\cite{chi2020fast} even though MACs from the FFT and IFFT are not counted. Our methods obtain comparable accuracy to regular ResNets with significantly reduced parameters and MACs. }
    \label{tab: ImageNet-1K}
\end{table*}

\begin{table*}[htbp]
    \centering
    \begin{tabular}{lcccc}
    \hline\noalign{\smallskip}
        \bf{Method}&\bf{Parameters (M)}&\bf{MACs (G)}&\bf{10-Crop Top1 Acc.}&\bf{Top5 Acc.}\\
        \noalign{\smallskip}\hline\noalign{\smallskip}
        %ResNet-18 (our trial, baseline)&11,689,512&72.05\%&90.81\%\\
        ResNet-18 (Torchvision~\cite{torchvision}, baseline)&11.69&1.822&71.86\%&90.60\%\\
        \bf{DCT-ResNet-18}&\bf{6.14 (47.5\%$\downarrow$)}&\bf{1.374 (24.6\%$\downarrow$)}&\bf{70.09\%}&\bf{89.40\%}\\
        \bf{Tripod-DCT-ResNet-18}&\bf{7.56 (35.3\%$\downarrow$)}&\bf{1.377 (24.4\%$\downarrow$)}&\bf{71.91\%}&\bf{90.55\%}\\
         \bf{Tripod-DCT-ResNet-18+1DCT-P}&\bf{7.82 (33.1\%$\downarrow$)}&\bf{1.378 (24.4\%$\downarrow$)}&\bf{72.20\%}&\bf{90.51\%}\\
        %ResNet-18+1DCT-Perceptron&11,952,778 (2.25\%$\uparrow$)&72.89\% (0.44\%$\uparrow$)&90.98\% (0.17\%$\uparrow$)\\
        \hline
        ResNet-50 \cite{he2016deep} (official~\cite{DeepResidualNetworksOfficialCode})&25.56&3.86&77.1\%&93.3\%\\
        ResNet-50 \cite{he2016deep} (Torchvision~\cite{torchvision})&25.56&4.122&77.43\%&93.75\%\\
        ResNet-50 (our trial, baseline)&25.56&4.122&77.53\%&93.75\%\\
        \bf{DCT-ResNet-50}&\bf{18.28 (28.5\%$\downarrow$)}&\bf{2.772 (32.8\%$\downarrow$)}&\bf{76.75\%}&\bf{93.44\%}\\
        \bf{Tripod-DCT-ResNet-50}&\bf{20.15 (21.1\%$\downarrow$)}&\bf{2.780 (32.6\%$\downarrow$)}&\bf{77.36\%}&\bf{93.62\%}\\
        \bf{Tripod-DCT-ResNet-50+1DCT-P}&\bf{24.35 (4.7\%$\downarrow$)}&\bf{2.783 (32.5\%$\downarrow$)}&\bf{77.41\%}&\bf{93.66\%}\\
        %ResNet-50+1DCT-Perceptron&29,755,530 (16.43\%$\uparrow$)&77.67\% (0.26\%$\uparrow$)&93.84\% (0.09\%$\uparrow$)\\
        %ResNet-50+1DCT-Px3&38,144,334 (49.25\%$\uparrow$)&\% (\%$\uparrow$)&\% (\%$\uparrow$)\\
        \noalign{\smallskip}\hline\noalign{\smallskip}
    \end{tabular}\vspace{-10pt}
    \caption{ImageNet-1K 10-crop results. Our models obtain comparable accuracy with significantly reduced parameters and MACs.}
    \label{tab: ImageNet-1K-10}
\end{table*}

\begin{table*}[htbp]
    \centering
    \begin{tabular}{lcccccc}
    \hline\noalign{\smallskip}
        \bf{Method}&\bf{Parameters (M)}&\bf{MACs (G)}&\bf{Center-Crop Top-1}&\bf{Top-5}&\bf{10-Crop Top-1}&\bf{Top-5}\\
        \noalign{\smallskip}\hline\noalign{\smallskip}
         ResNet-18&11.69&1.822&69.76\%&89.08\%&71.86\%&90.60\%\\
         %ResNet-18&11,689,512&69.99\%&89.45\%&72.05\%&90.81\%\\
        \bf{ResNet-18+1DCT-P}&\bf{11.95}&\bf{1.823}&\bf{70.50\%}&\bf{89.29\%}&\bf{72.89\%}&\bf{90.98\%}\\
        %\hline
        % \noalign{\smallskip}\hline\noalign{\smallskip}
        ResNet-50&25.56&4.122&76.06\%&92.85\%&77.53\%&93.75\%\\
        \bf{ResNet-50+1DCT-P}&\bf{29.76}&\bf{4.124}&\bf{76.09\%}&\bf{93.80\%}&\bf{77.67\%}&\bf{93.84\%}\\
        %ResNet-50+1DCT-Px3&38,144,334(49.25\%$\uparrow$)&\% (\%$\uparrow$)&\% (\%$\uparrow$)\\
        % +1DCT-MLP&50,727,148 (\%$\uparrow$)&\% (\%$\uparrow$)&\% (\%$\uparrow$)&\\
        % \bf{DCT-ResNet-50 MLP}&\bf{24,707,814 (\%$\downarrow$)}&\bf{\% (\%$\downarrow$)}&\bf{\% (\%$\downarrow$)}\\
        \noalign{\smallskip}\hline\noalign{\smallskip}
    \end{tabular}\vspace{-10pt}
    \caption{ImageNet-1K results of inserting an additional DCT-perceptron layer before the global average pooling layer. The additional DCT-perceptron layer improves the accuracy of ResNets without changing the main structure of the ResNets.}
    \label{tab: ImageNet-1K extra}
\end{table*}

\subsection{Ablation Study: CIFAR-10 Classification}
Training ResNet-20 and DCT-ResNet-20s follows the implementation in~\cite{he2016deep}. We use an SGD optimizer with a weight decay of 0.0001 and momentum of 0.9. These models are trained with a mini-batch size of 128. The initial learning rate is 0.1 for 200 epochs, and the learning rate is reduced by a factor of 1/10 at epochs 82, 122, and 163, respectively. Data augmentation is implemented as follows: First, we pad 4 pixels on the training images. Then, we apply random cropping to get 32 by 32 images. Finally, we randomly flip images horizontally. We normalize the images with the means of [0.4914, 0.4822, 0.4465] and the standard variations of [0.2023, 0.1994, 02010]. During the training, the best models are saved based on the accuracy of the CIFAR-10 test dataset, and their accuracy numbers are reported in Table~\ref{tab: CIFAR-10}. 

\begin{figure}[htbp]
\centering
\subfloat{\includegraphics[width=0.45\linewidth]{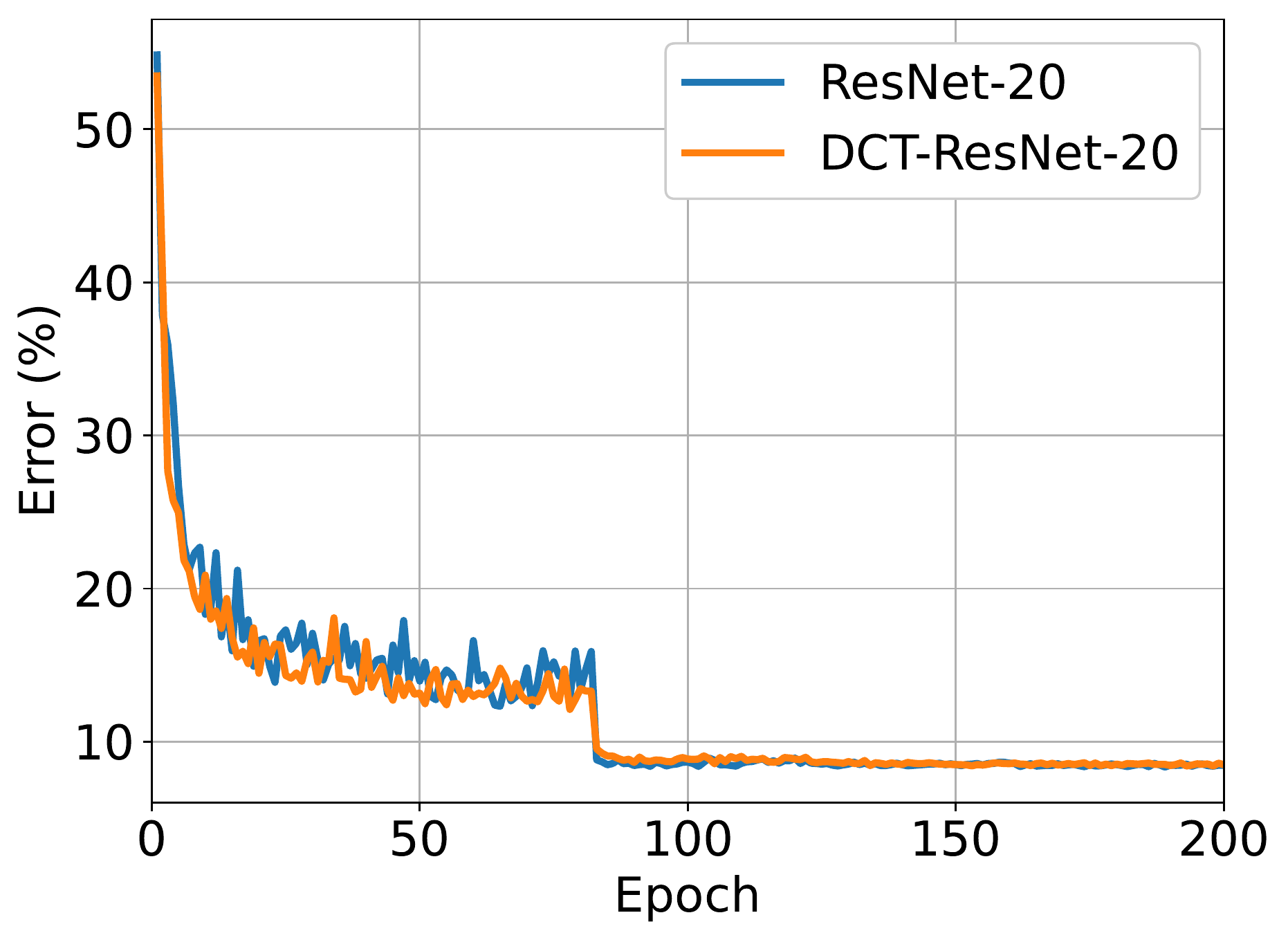}}
\subfloat{\includegraphics[width=0.45\linewidth]{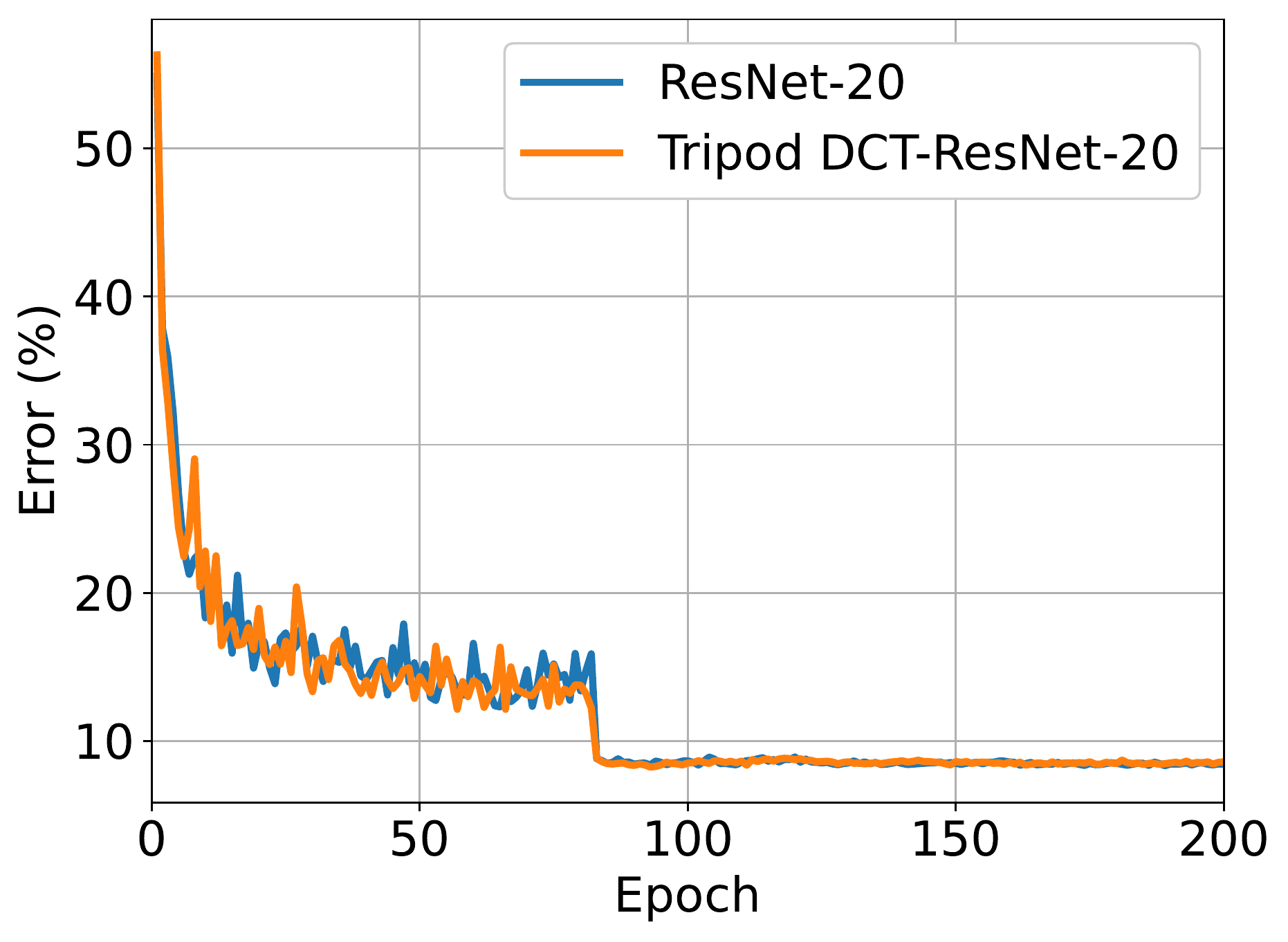}}\vspace{-5pt}
\caption{Training on CIFAR-10. Curves denote test error. Left: ResNet-20 vs DCT-ResNet-20. Right: ResNet-20 vs Tripod DCT-ResNet-20. More than 26\% parameters are reduced using the DCT-perceptron layers, while accuracy results are comparable.\vspace{-5pt}}
\label{fig: CIFAR-10}
\end{figure}

Figure \ref{fig: CIFAR-10} shows the CIFAR-10 test error history during the training. As shown in Table~\ref{tab: CIFAR-10}, the single-pod DCT-perceptron layer-based ResNet (DCT-ResNet-20) has 44.39\% lower parameters than regular ResNet-20, but the model only suffers an accuracy loss of 0.07\%. On the other hand, the tripod DCT-perceptron layer-based ResNet (TriPod-DCT-ResNet-20) achieves 0.09\% higher accuracy than the baseline ResNet-20 model. TriPod-DCT-ResNet-20 has 26.64\% lower parameters than regular ResNet-20. When we add an extra DCT-perceptron layer to the regular ResNet-20 (ResNet-20+1DCT-P) we even get a higher accuracy (91.82\%) as shown in Table~\ref{tab: CIFAR-10}.

In the ablation study on the DCT-perceptron layer, first, we first remove the residual design (the shortcut connection) in the DCT-perceptron layers and get a worse accuracy (91.12\%). Then, we remove scaling which is convolutional filtering in the DCT-perceptron layers. In this case, the multiplication operations in the DCT-perceptron layer are only implemented in the $1\times 1$ Conv2D. The accuracy drops from 91.59\% to 90.46\%. Therefore, scaling is necessary to maintain accuracy. Next, we apply ReLU instead of soft-thresholding in the DCT-perceptron layers. We first apply the same thresholds as the soft-thresholding thresholds on ReLU. In this case, the number of parameters is the same as in the proposed DCT-ResNet-20 model, but the accuracy drops to 91.30\%. We then try regular ReLU with a bias term in the $1\times 1$ Conv2D layers in the DCT-perceptron layers, and the accuracy drops further to 91.06\%. These experiments show that the soft-thresholding is superior to the ReLU in the DCT analysis, as the soft-thresholding retains negative DCT components with high amplitudes, which are also used in denoising and image coding applications. Furthermore, we replace all Conv2D layers in the ResNet-20 model with the DCT-perceptron layers. We implement the downsampling (Conv2D with the stride of 2) by truncating the IDCT2D so that each dimension of the reconstructed image is one-half the length of the original. In this method, 81.31\% parameters are reduced, but the accuracy drops to 85.74\% due to lacking parameters. 

In the ablation study on the tripod DCT-perceptron layer, we also try the bipod, the quadpod, and the quintpod structures, but their accuracy is lower than the tripod. Moreover, we also try adding the residual design in the tripod DCT-perception layer, but our experiments show that the shortcut connection is redundant. This may be because there are three channels in the tripod structure replacing the shortcut. The multiple-channeled structure allows the derivatives to propagate to earlier channels better than a single channel. %Furthermore, we try to replace all Conv2D layers as the ablation study on the single-pod DCT-perceptron layer, and the model suffers 4.96\% accuracy loss because of also lacking parameters. On the other hand, if we retain some Conv2D layers (There are still more Conv2D layers replaced than in the proposed tripod-DCT-ResNet-20), the accuracy dropping is reduced. Therefore, reducing too many parameters is not an optimal choice for such a task, as accuracy also drops much.

\subsection{ImageNet-1K Classification}
We employ PyTorch's official ImageNet-1K training code~\cite{ImageNet_training_in_PyTorch} in this section. Since we use the PyTorch official training code with the default training setting to train the revised ResNet-18s, we use the official trained ResNet-18 model from the PyTorch Torchvision as the baseline network. We use an NVIDIA RTX3090 to train the ResNet-50 and the corresponding DCT-perceptron versions. The default training needs about 26--30 GB GPU memory, while an RTX3090 only has 24GB. Therefore, we halve the batch size and the learning rate correspondingly.  We use an SGD optimizer with a weight decay of 0.0001 and momentum of 0.9. DCT-perceptron-based ResNet-18's are trained with the default setting: a mini-batch size of 256, an initial learning rate of 0.1 for 90 epochs. ResNet-50 and DCT-Perceptron-based models are trained with a mini-batch size of 128, the initial learning rate is 0.05 for 90 epochs. The learning rate is reduced by a factor of 1/10 after every 30 epochs. For data argumentation, we apply random resized crops on training images to get 224 by 224 images, then we randomly flip images horizontally. We normalize the images with the means of [0.485, 0.456, 0.406] and the standard variations of [0.229, 0.224, 0.225], respectively. We evaluate our models on the ImageNet-1K validation dataset and compare them with the state-of-art papers. During the training, the best models are saved based on the center-crop top-1 accuracy on the ImageNet-1K validation dataset, and their accuracy numbers are reported in Tables~\ref{tab: ImageNet-1K} and \ref{tab: ImageNet-1K-10}.

\begin{figure}[t]
\centering
\subfloat{\includegraphics[width=0.45\linewidth]{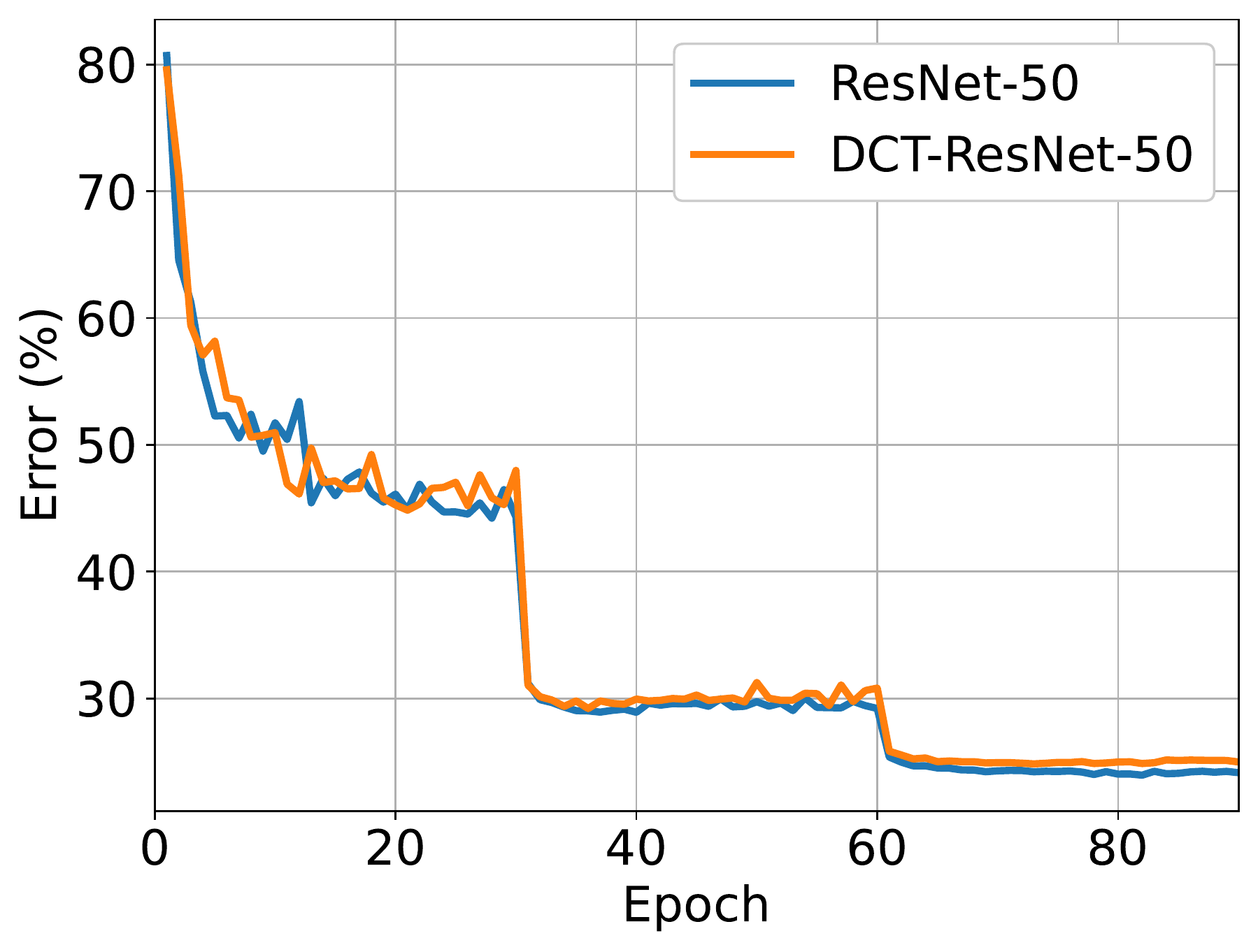}}
\subfloat{\includegraphics[width=0.45\linewidth]{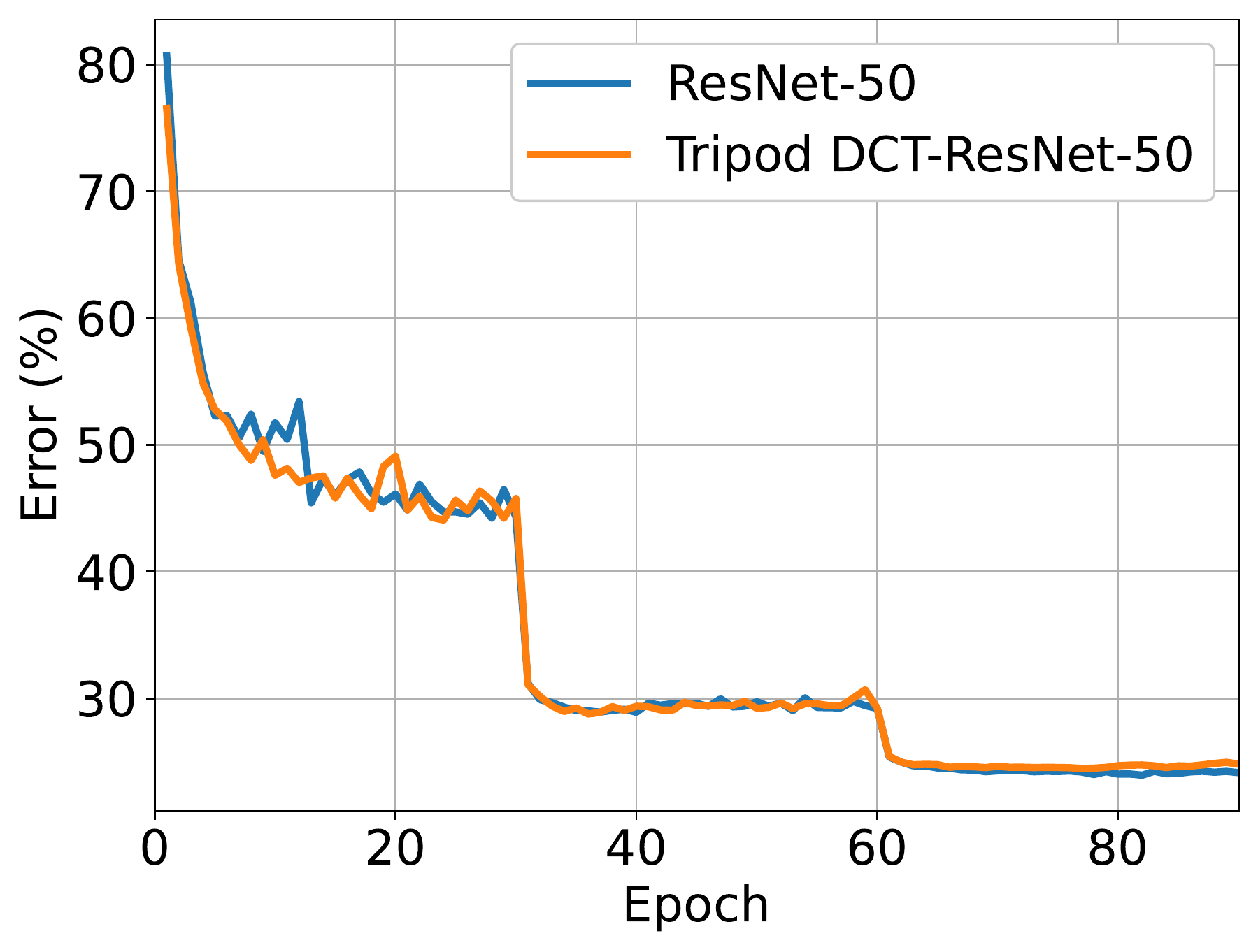}}\vspace{-5pt}
\caption{Training on ImageNet-1K. Curves denote the validation error of the center crops. Left: ResNet-50 vs DCT-ResNet-50. Right: ResNet-50 vs Tripod DCT-ResNet-50. More than 21\% parameters and 32\% MACs are reduced using the DCT-perceptron layers with comparable accuracy to regular ResNet-50.\vspace{-5pt}}
\label{fig: ImageNet-1K}
\end{figure}

Figure \ref{fig: ImageNet-1K} shows the test error history on the ImageNet-1K validation dataset during the training phase. As shown in Table~\ref{tab: ImageNet-1K},  we reduce 35.3\% parameters and 22.4\% MACs of regular ResNet-18 using the tripod DCT-perceptron layer, and the center-crop top-1 accuracy only drops from 69.76\% to 69.55\%. It improves the 10-Crop top-1 accuracy from 71.86\% to 71.91\%. The single-pod DCT-ResNet-18 achieves a relatively poor accuracy (67.84\%) because the parameters are insufficient (47.5\% reduced) for the ImageNet-1K Task. The DCT-ResNet-50 model contains 28.5\% less parameters with 32.8\% less MACs than the baseline ResNet-50 model, and its center-crop top-1 accuracy only drops from 76.06\% to 75.17\%. The tripod DCT-ResNet-50 model contains 21.1\% less parameters with 32.6\% less MACs than the baseline ResNet-50 model, while its center-crop top-1 accuracy only drops from 76.06\% to 75.52\%.

We can also increase the accuracy of regular ResNets by inserting an additional DCT-perceptron layer with batch normalization after the global average pooling layer of ResNets. We call these models as ResNets+1DCT-P. This additional layer leads to a negligible increase in MACs. As shown in Table~\ref{tab: CIFAR-10}, this additional layer increases the accuracy of ResNet-20 on CIFAR-10 from 91.66\% to 91.82\% with only 1.60\% extra parameters. An additional DCT-Perceptron  layer increases the center-crop top-1 accuracy on ResNet-18 from 69.76\% to 70.50\% only with 2.3\% extra parameters with 0.05\% higher MACs, and it increases ResNet-50's center-crop top-5 accuracy from 92.85\% to 93.80\% with 16.4\% extra parameters with 0.05\% higher MACs. The additional layer can also be applied to the Tripod-DCT-ResNets to further improve the accuracy.

\section{Conclusion}
In this paper, we proposed a novel layer based on DCT to replace $3\times 3$ Conv2D layers in convolutional neural networks. The proposed layer is derived from the DCT convolution theorem. Processing in the DCT domain consists of a scaling layer, which is equivalent to convolutional filtering, a $1\times 1$ Conv2D layer, and a trainable soft-thresholding function. In our CIFAR-10 experiments, we replaced the $3\times 3$ Conv2D layers of ResNet-20 using the proposed DCT-perceptron layer. We reduced 44.39\% parameters by obtaining a comparable result (accuracy -0.07\%), or 26.64\% parameters by obtaining even a slightly better result (accuracy + 0.09\%), compared to the baseline ResNet-20 model. In our ImageNet-1K experiments, we also obtain comparable results (accuracy drops less than 0.8\%) compared to regular ResNet-50 while reducing the number of parameters by more than 21\% and MACs by more than 32\%.
Furthermore, we improve the results of regular ResNets with an additional layer of DCT-perceptron. We insert the proposed DCT-perceptron layer into regular ResNets with batch normalization before the global average pooling layer to improve the accuracy with a slight increase of MACs.

%%%%%%%%% REFERENCES
{\small
\bibliographystyle{unsrt}
\bibliography{main}
}

\end{document}